\documentclass[letterpaper, 10 pt, conference]{packages/ieeeconf}  

\IEEEoverridecommandlockouts                              

\usepackage{times}

\usepackage{acronym}
\usepackage{amsmath}
\usepackage{array}
\usepackage{graphicx} 
\usepackage{multicol}
\usepackage{subcaption}
\usepackage{tabularx}
\usepackage{verbatim}
\usepackage{xcolor}
\usepackage{flushend}

\usepackage[bookmarks=true]{hyperref}

\usepackage{packages/abbreviations}

\acrodef{cnn}[CNN]{Convolutional Neural Network}
\acrodef{gan}[GAN]{Generative Adversarial Network}
\acrodef{lstm}[LSTM]{Long Short-Term Memory}
\acrodef{sgan}[S-GAN]{Social GAN}

\title{\LARGE \bf
Deep Context Maps: Agent Trajectory Prediction\\ using Location-specific Latent Maps
}

\author{Igor Gilitschenski$^{1}$,  Guy Rosman$^{2}$, Arjun Gupta$^{3}$,
Sertac Karaman$^{3}$, Daniela Rus$^{1}$
\thanks{*This work has been supported by the Toyota Research Institute. It, however, reflects solely the opinions and conclusions of its authors and not TRI or any other Toyota entity. The support is
gratefully acknowledged.}
\thanks{$^{1}$MIT Computer Science and Artificial Intelligence Lab (CSAIL), {\tt\small igilitschenski@mit.edu}, {\tt\small rus@csail.mit.edu}}%
\thanks{$^{2}$Toyota Research Institute, {\tt\small guy.rosman@tri.global}}
\thanks{$^{3}$MIT Laboratory for Information and Decision Systems (LIDS), {\tt\small argupta@mit.edu}, {\tt\small sertac@mit.edu}}%
}

\begin{document}

\maketitle

\begin{abstract}
In this paper, we propose a novel approach for agent motion prediction in cluttered environments. One of the main challenges in predicting agent motion is accounting for location and context-specific information. Our main contribution is the concept of learning context maps to improve the prediction task. Context maps are a set of location-specific latent maps that are trained alongside the predictor. Thus, the proposed maps are capable of capturing location context beyond visual context cues (e.g. usual average speeds and typical trajectories) or predefined map primitives (such as lanes and stop lines). We pose context map learning as a multi-task training problem and describe our map model and its incorporation into a state-of-the-art trajectory predictor. In extensive experiments, it is shown that use of learned maps can significantly improve predictor accuracy. Furthermore, the performance can be additionally boosted by providing partial knowledge of map semantics. 
\end{abstract}

\IEEEpeerreviewmaketitle

\section{INTRODUCTION}

%
%
Trajectory prediction of diverse agents in dynamic environments is a key challenge towards unlocking the full potential of autonomous mobile robots.  Particularly in safety critical settings such as autonomous driving, obtaining reliable predictions of surrounding agents is a necessary functionality for robust operation at speeds comparable to human-driven vehicles. Predicting agent trajectories, such as pedestrian motions, is an inherently challenging task:  First, in crowded environments, pedestrian behavior is highly dependent on social interactions. Second, prediction systems have to account for potential rapid changes in behaviour (e.g. children unexpectedly running on the road). Finally, the decision making process involves and depends on a high variety of factors including the surrounding environment and the complex interactions with it.

%
%
Several deep-learning based approaches have recently incorporated context to improve prediction accuracy~\cite{Lee2017,Kosaraju2019, Sadeghian2019}. This was usually achieved by providing a top-down view of the scenery to the prediction network \cite{Sadeghian2019}, or by adding information (e.g. about lanes) from a map of the area \cite{chang2019argoverse}. This improves prediction performance but fails to account for information that is not available through visual cues or the content of a map. Furthermore, existing map-based prediction approaches usually rely on a map creation process that requires additional engineering and reasoning about useful features for prediction. To the best of our knowledge, there is no prediction approach that implicitly learns a task-specific latent map.

%
%
We argue that prediction performance can be improved by revising the way predictors and maps interact. Typical local behavior is an important cue in predicting behaviors and affordances. This is strongly evident in autonomous vehicle fleets that regularly traverse the same roads. Maps should therefore extend beyond the content of a top-down image or an HD map with fixed semantics. They should provide predictors with location-specific context in a structured way. We propose learning a location-specific joint representation (called context map) that explains top-down views, semantic primitives, and the behavior of agents operating in the scene.

%
%
For instance, the ever-growing amount of raw trajectory data contains a lot of information about local norms. Trajectories also implicitly encode phenomena such as common paths, potential obstacles, and traffic flow. Thus, they could be a rich source of non-visual information. However, it is an open problem how to incorporate such information into a context representation for the prediction task. A related challenge is the decomposition of location-specific and location-agnostic learned structures within the predictor's architecture.


%
%
\begin{figure}[t]
    \centering
    \includegraphics[width=0.9\columnwidth]{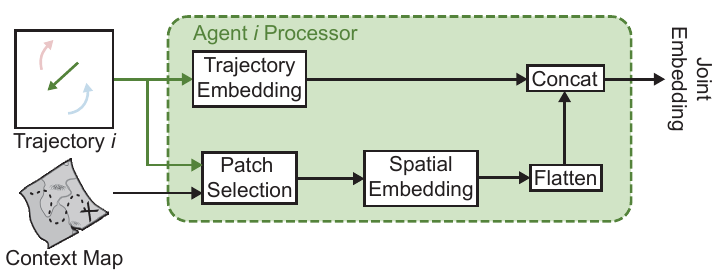}
    \caption{{ \bf Map Embedding.} Latent maps can be used for learning location-specific information. For the trajectory forecasting task, we preprocess the latent map and the trajectory resulting in a joint trajectory-map embedding.}
    \label{fig:processor}
\end{figure}

%
%
\begin{figure*}[t]
    \centering
    \includegraphics[width=0.9\textwidth]{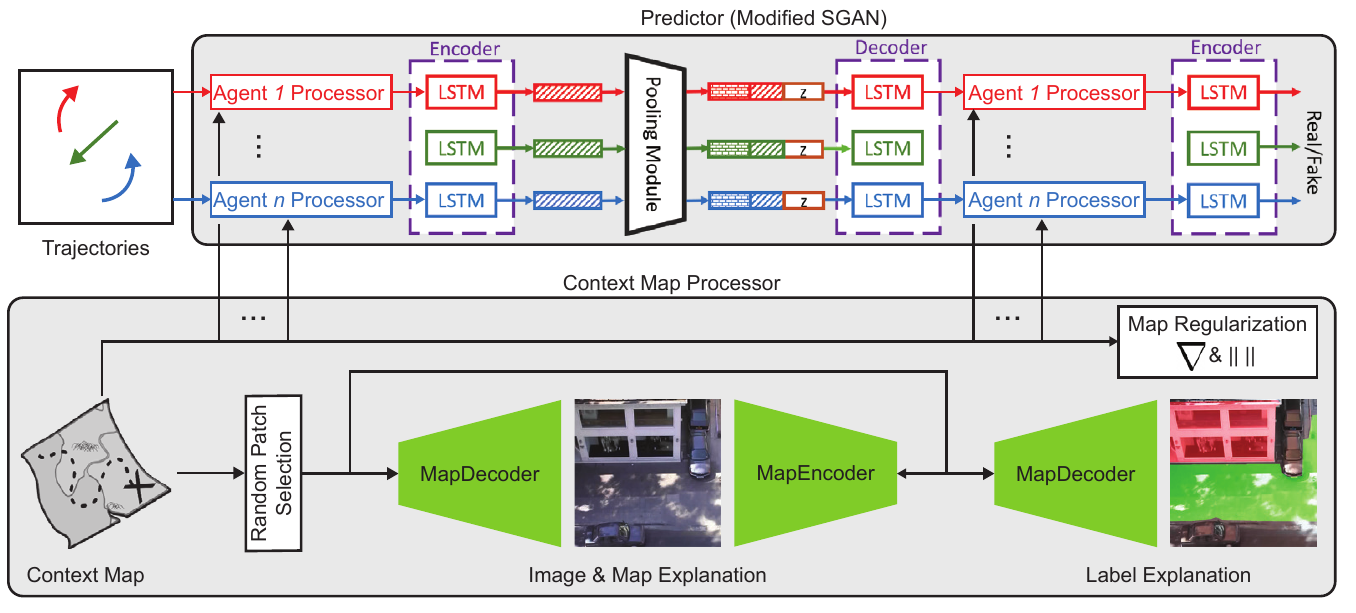}
    \caption{{ \bf Architecture Overview.} The proposed overall architecture for learning context maps jointly with trajectory prediction. For the prediction network, we use a modified variant of the Social GAN architecture (\cite[Fig.2]{Gupta2018}) that introduces the context map after the trajectory embedding \cite[Eqns. 2 \& 4]{Gupta2018}. Additional supervision for learning the context map is provided image explanation, map explanation, and label explanation losses on randomly sampled map patches and the corresponding image patches. Finally, the map is regularized using a norm penalization loss, and a gradient norm penalization loss to enforce sparsity of the map gradient.}
    \label{fig:architecture}
\end{figure*}

%
%
In this paper, we address these challenges by proposing an approach for neural network-based prediction that incorporates context maps, a learned location-specific memory.  Instead of presuming a predetermined structure, we train the maps as latent entities that can not only explain visual features of the image, but also non-visual features for the prediction task. The trained maps are implemented as a set of location-specific biases that are injected into the prediction network. Additional auxiliary loss terms based on reconstruction, partial semantic annotations, and gradient sparsity provide support to guide the map-learning process. At test time, the network uses current locations of the agents and the learned maps as input for prediction. To the best of our knowledge, this is the first work focusing explicitly on latent map learning for improving the prediction task. Overall, our contributions can be summarized as follows:
\begin{itemize}
\item We develop a model for utilizing latent maps to explain trajectories of road agents and integrate them as part of a state-of-the-art trajectory prediction network.  
\item We show how we learn the maps from raw aerial imagery as well as the observed motion patterns and partial semantic labels. 
\item We demonstrate how the learned maps allow us to better predict agents' motion and outperform baseline approaches for the prediction task. We show this on standard benchmark datasets and probe the performance contribution of the maps and additional semantic labels.
\end{itemize}

\section{RELATED WORK}
Trajectory modeling and prediction has been covered in an extensive and diverse body of work. Early works consider prediction in the context of tracking~\cite{koller1993model} or social interaction modelling~\cite{Helbing1995}. Recent advances in deployment of autonomous vehicles have sparked renewed interest in the prediction task (see \cite{lefevre2014survey,Rudenko2019,Ridel2018} for recent surveys). A main challenge remains the incorporation of environmental context.

\textbf{Social-Context Modelling} One big line of work, inspired by~\cite{Helbing1995}, considers improving prediction by properly modelling social context and group dynamics. In~\cite{Vemula2017}, future trajectories of all interacting agents are modeled by learning social interactions from real data using a Gaussian process model. An interaction-aware prediction network is used in \cite{Radwan2018} for safely crossing an intersection. In~\cite{Pfeiffer2018}, static obstacles and surrounding pedestrians are explicitly modeled for improving the forecasting task while~\cite{Ivanovic2019} proposes an approach that uses graphs for scene representation. Social Pooling modules are proposed in \emph{Social LSTM}~\cite{Alahi2016} and \emph{Social GAN}~\cite{Gupta2018}. They allow a deep learning-based predictor to jointly reason about multiple agent trajectories.  In contrast to these approaches, the present work considers the orthogonal problem of semantic context modelling and can be combined with modelling social context as we demonstrate by integrating Context Maps with \emph{Social GAN}. We show that proper modelling of semantic context has a stronger impact on prediction accuracy than social pooling. 

%
%

%
%
\textbf{Semantic Context Modelling}. Several recent approaches consider broader semantic context information for trajectory prediction. In~\cite{Wiest2012}, previously observed motion patterns are used to estimate a probability distribution as motion model and \cite{Coscia2018} estimates circular distributions at different locations which are combined into a smooth path prediction. An existing map with annotated traffic lanes and centerlines is used for prediction in~\cite{Petrich2013} based on a Kalman Filter framework for predicting vehicle motion. The work \cite{Ballan2016} proposes to extract patch descriptors that encode the probability of moving to adjacent patches and then uses a Dynamic Bayesian Network for scene prediction. In~\cite{Habibi2018}, context features from the environment (such as traffic light status and distance to curbside) are incorporated into a Gaussian Process-based predictor. Several deep learning-based approaches use visual context cues for improving on the prediction task~\cite{Lee2017, Kooij2018, Kosaraju2019, Sadeghian2018, Xue2018, Sadeghian2019} by taking an image of the scenery as an additional input to a network. Different map representations for prediction in scenarios with static trackers are discussed in~\cite{Kitani2012, Ridel2020, Jacobs2017, Rudenko2018}. An explicit destination network is used in~\cite{Rehder2018} to model a grid of potential destinations for subsequent trajectory prediction. An implicit consideration of semantic information is achieved in networks that combine multiple tasks such as~\cite{Luo2018}, where a single convolutional network is used to combine detection, tracking, and motion prediction. In contrast to these works, our work extracts location-specific semantic information  during training and stores it in a learned latent representation. We also use different information sources to inform that representation. 

%
%
\textbf{Latent Representation Learning.} Our work draws some inspiration from several approaches that encode learned map representations. An early approach of map learning was presented in~\cite{Ramos2016}, where a Gaussian Process is used for occupancy mapping without a priori discretization of the world into grid cells. In~\cite{Gupta2017}, a differentiable mapper is used to create a multiscale belief of the world in the agent's coordinate frame. Other recent neural mapping approaches involve~\cite{Parisotto2017, Zhang2017, Avraham2019}. While these works mostly focus on localization and navigation, we use a latent location-specific memory to inform the prediction task.

%
%
The most similar work to ours is the predictor proposed in~\cite{Yi2016}. This approach uses a displacement volume as a network input and proposes a location-specific bias map of the size of that volume.  In contrast to this work, we consider a model capable of simultaneously handling multiple scenes and multiple map layers. We also propose a location-specific training methodology and additional semantic supervision to improve latent map quality.

\begin{figure}[t]
    \centering
    \includegraphics[width=0.9\columnwidth]{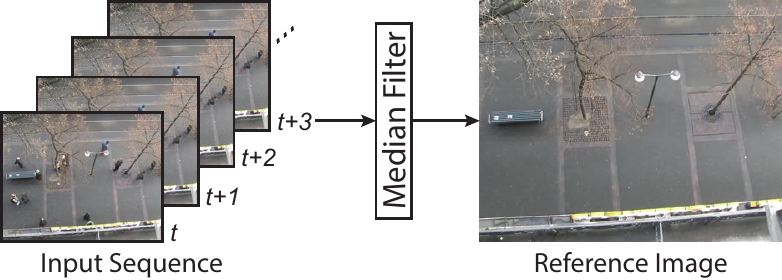}
    \caption{{ \bf Reference Images.} We create reference images of the scenes by applying a median filter. This results in the removal of most dynamic objects and supports scene reconstruction.}
    \label{fig:medianfilter}
\end{figure}

\section{CONTEXT MAP LEARNING FOR PREDICTION}
The goal of the proposed approach is to capture the ability of humans to account for environment and location-specific habits and norms. We model this information by a set of location-specific maps that are learned during predictor training. 

\begin{figure}
	\centering
	\begin{subfigure}[t]{0.2\textwidth}
		\centering
		\includegraphics[width=\textwidth]{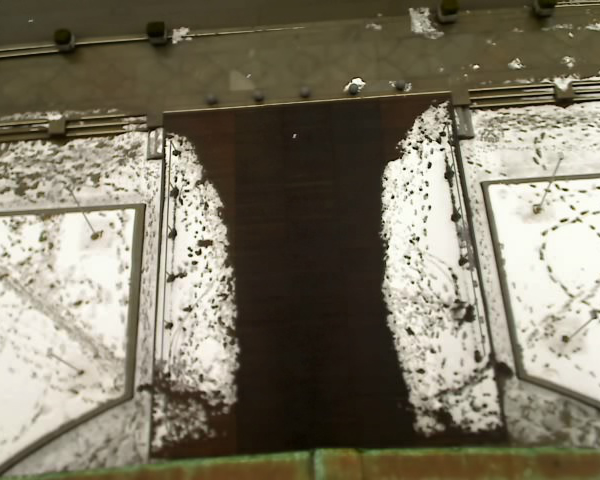}
		\caption{ETH}
	\end{subfigure}%
	~ 
	\begin{subfigure}[t]{0.2\textwidth}
		\centering
		\includegraphics[width=\textwidth]{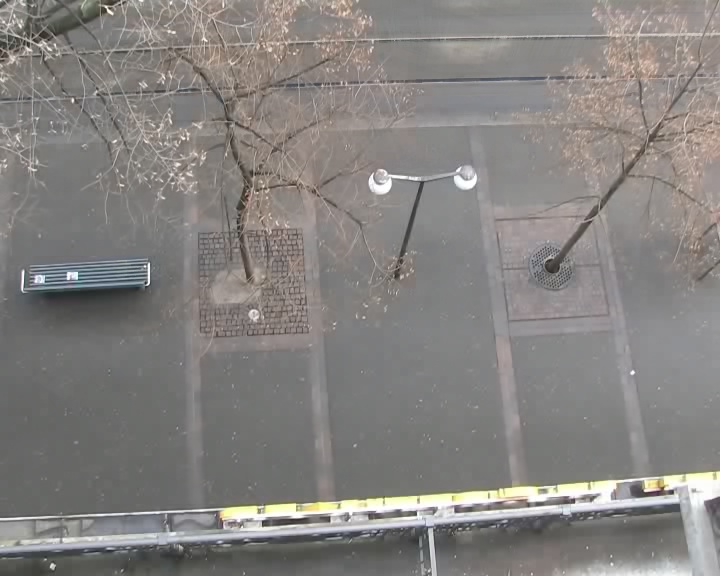}
		\caption{Hotel}
	\end{subfigure}
	\begin{subfigure}[t]{0.2\textwidth}
		\centering
		\includegraphics[width=\textwidth]{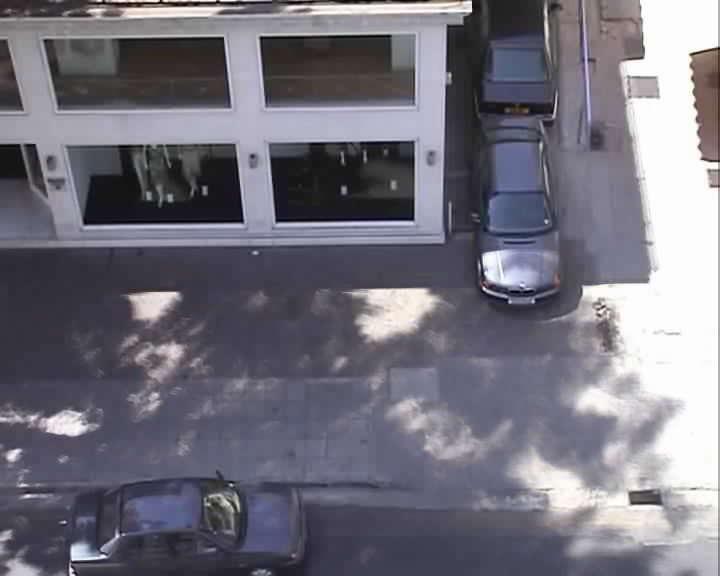}
		\caption{Zara1}
	\end{subfigure}%
	~ 
	\begin{subfigure}[t]{0.2\textwidth}
		\centering
		\includegraphics[width=\textwidth]{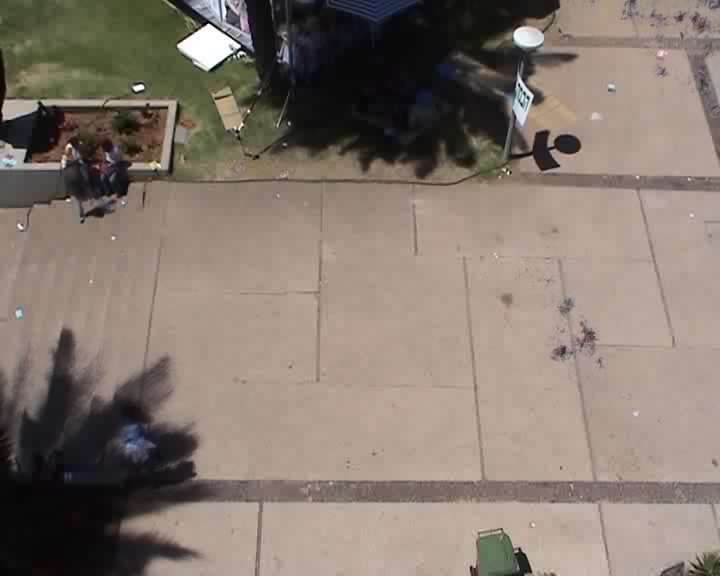}
		\caption{Students3}
	\end{subfigure}
	
	\caption{{\bf Reference Images.} The reference images that are used as supervisory reconstruction labels for our networks are generated by applying a temporal median filter \cite{Benezeth2008} to the video data in the datasets.}\label{fig:reference}
\end{figure}

More formally, our goal is to predict a set of agent trajectories $\fhY=\{Y_1, ... Y_N\}$ at a place $l\in\mcP$ (with $\mcP$ denoting the set of all considered places) from past temporally overlapping trajectories $\fX=\{X_1, \ldots, X_N\}$ and a learned location-specific map $M_l$, i.e.
\[
\fhY = f\li(\fX, M_l\ri)\ .
\]
such that $\fhY$ approximates the ground truth trajectories, $\fY$, as closely as possible. The trajectories are represented as sequences $X_i=\{\fx_{i,t}\in\R^2\,|\, t=1,\ldots, O\}$ and $Y_i=\{\fy_{i,t}\in\R^2\,|\, t=1,\ldots, P\}$ with observation horizon $O$ and prediction horizon $P$. The neural network representing $f$ is trained together with the context maps $M_l$. The maps are stored as tensors of size $H_l \times W_l \times F_{map}$ with $H_l$, $W_l$ denoting the reference image dimensions and $F_{map}$ the map feature dimension. 
In our case, reference images are usually top-down views of the environment obtained from the video data of the considered datasets via median filtering as shown in Fig.~\ref{fig:medianfilter}. The resulting reference images are visualized in Fig.~\ref{fig:reference}.

In addition to trajectory losses during predictor training (Sec.~\ref{sec:predictor}), we provide weak supervisory information to obtain meaningful maps and ensure convergence. We train the maps to reconstruct the reference image (Sec~\ref{sec:reconstruction}), to encode information about environment semantics without providing full labels on the entire reference image (Sec~\ref{sec:semantics}), and add a gradient based penalty term to support map smoothness (Sec~\ref{sec:sparsity}).

\subsection{Predictor Integration}\label{sec:predictor}

Recently several generative trajectory prediction approaches based on \acp{gan}~\cite{Goodfellow2014} have been proposed demonstrating the capability for covering a variety of different plausible trajectories~\cite{Gupta2018,Sadeghian2018,Kosaraju2019}. 
Motivated by these results and in order to prove usefulness of maps even with elaborate predictors, we integrated our map learning approach with \ac{sgan}~\cite{Gupta2018} as visualized in Fig.~\ref{fig:architecture}. The concept of context map learning is applicable to most neural network based predictors. We used \ac{sgan} due to the free availability of its implementation allowing for a fair baseline comparison.

The \ac{sgan} generator network creates trajectory predictions through a \ac{lstm}~\cite{Hochreiter1997} network which is broadly used by several of the above-mentioned trajectory prediction models. Traditionally, for trajectory prediction, the \ac{lstm} network takes in a sequence of agent coordinates, encodes them into a state vector, and a separate predictor network converts the state vector to the future agent location. The \ac{sgan} network simultaneously processes the trajectories of all the pedestrians in a given consecutive sequence of video frames and then ``pools" the resulting state vectors of the separate \acp{lstm} before making a prediction. The pooling mechanism serves for modeling social interactions. More formally, for each trajectory $X_i$, the \ac{sgan} \ac{lstm} cell follows the following recurrence:
\begin{equation}\label{eq:generatorLSTM}
\begin{split}
\fe_c &= \mathrm{MLP}(\fx_{i,t})\ ,\\
\fh_t
 &= \mathrm{LSTM}(\fh_{t-1},\, \fe_{c})\ ,
\end{split}
\end{equation}
where MLP denotes a multi layer perceptron meant to encode the coordinates of the agent, and $\fh_t$ is the hidden state of the \ac{lstm} at time $t$. This computation is carried our for each trajectory in $\fX$ individually, however for simplicity of notation, we do not carry the index of the trajectory as it is clear for the context.

As the model is a GAN, it also includes a  discriminator network which scores the trajectory produced by the generator. This network is only used during training to improve the generator and not part of the trajectory prediction network at inference time. 

We integrate context maps with the \ac{sgan} model by providing an additional input during the prediction phase to the \ac{lstm}. We add a \ac{cnn} that takes in a patch of the context map for the given scene at the current coordinate location and provides a processed form to the first \ac{lstm} cell in the generator. This additional input changes the recurrence in \eqref{eq:generatorLSTM} to 
\begin{equation}
\begin{aligned}
\fe_m &= \mathrm{MapDecoder_P}(C_{i,t})\, , &
\fe_t &= \mathrm{MLP}(\fx_{i,t})\, ,\\
\fe_c &= \mathrm{Concat}(\fe_t, \fe_m)\, , &
\fh_t &= \mathrm{LSTM}(\mathbf{h}_{t-1}, \fe_{c})\, ,
\end{aligned}
\end{equation}
where $\mathrm{MapDecoder_P}(\cdot)$ is a Convolutional Neural Network creating a spatial embedding of the map and $C_{i,t}$ is a patch of the context map around the location $\fx_{i,t}$. 
This process is visualized in Fig.~\ref{fig:processor}. To make a prediction on the future location of the agent, we pass the most recent \ac{lstm} state vector to a fully connected decoder network which outputs the future position of the agent (analogous to \cite[eq. (4)]{Gupta2018}).

At training time, the generator involves two loss terms. In addition to the usual discriminator score $\mcL_{score}$, \ac{sgan} uses a L2 loss term between the predicted trajectory and the true trajectory
\begin{equation}
\mcL_{\mathrm{traj}}(\fhY, \fY) = \sum_{i=1}^N \sum_{t=1}^P \norm{\fhy_{i,t} - \fy_{i,t}}^2\ .
\end{equation}
In our modified version, the discriminator also gets access to the maps using the same augmentation process as for the observed trajectories. However, in contrast to generator training, the latent map is not modified during discriminator training iterations.
 
\subsection{Map Patch Selection}
One of the conceptual choices for the predictor was deciding between using global context of the entire scene (as is done e.g. in~\cite{Sadeghian2018}) or a local context notion. We decided for the latter to simplify the learning task.  To ensure local use of context, the network is implemented such that during each training and inference step only those parts of the map $M_l$ are used that correspond to locations in the current batch or, for training only, to patches selected for auxiliary tasks. 
This requires a patch selection mechanism which is simply extracting subtensors of the map around a given point $\fx$ of size $H_{patch} \times W_{patch} \times F_{map}$ denoted as
\[
C = \mathrm{PatchSelection}(M_l, \fx)\ .
\]
Thus, $C_{i,t}$ above is obtained as $C_{i,t}=\mathrm{PatchSelection}(M_l, \fx_{i,t})$. 

\subsection{Image Explanation}\label{sec:reconstruction}
Unless the size of the training set for a specific environment is very large, the trajectories will usually not cover all walkable areas and, on their own, do not provide sufficient information about other objects in the environment. To help the network better learn key features of the environment, we introduce an image explanation mechanism as well as a map explanation mechanism.  We enforce this constraint by including decoder and encoder modules according to
\begin{equation}
\begin{aligned}
\mcL_{\mathrm{image}}(l) =
 & \norm{I_l - \mathrm{MapDecoder_R}(M_l)}\\
 & +\norm{M_l - \mathrm{MapEncoder_R}(I_l)}\ .
\end{aligned}
\end{equation}
We note that this methodology can potentially allow to use this prediction approach in unseen scenes.

\subsection{Semantic Label Reconstruction}\label{sec:semantics}
We expect the latent map to decode semantic labels of the scene where the annotation is present. To that end, we add another module to decode the context map into semantic class labels represented as a matrix $L_{l,i}\in \R^{H_l \times W_l}$ for each location $l$ and label type $i$. For positive and negative examples we set the values of $L_{l,i}$ to 1 and -1 respectively and leave it at 0 if there is no label information. The label reconstruction loss is formulated as the difference between the hand-annotated semantic labels and the decoded annotation
\[
\mcL_{\mathrm{labels}}(l) = \sum_{i\in \mcT}(L_{l,i} - B_{l,i} \circ \mathrm{MapDecoder}_S(M_l)_i)^2\ ,
\]
where $\mcT$ denotes the set representing label types, $\circ$ is the Hadamard product and $B_{l,i}$ is a bitmask ensuring that no loss is incurred for areas without any label (i.e. it is 0 wherever $L_{l,i}$ is 0 and 1 otherwise).

\begin{table*}[t]
\centering
\begin{tabular}{c|ccccc}
{\bf Category} & 
{\bf ETH} & 
{\bf HOTEL} & 
{\bf STUDENTS3} & 
{\bf ZARA1} & 
{\bf ZARA2}
\\ \hline 
\raisebox{0.8cm}{\bf Walkable} & 
\includegraphics[width=0.3\columnwidth]{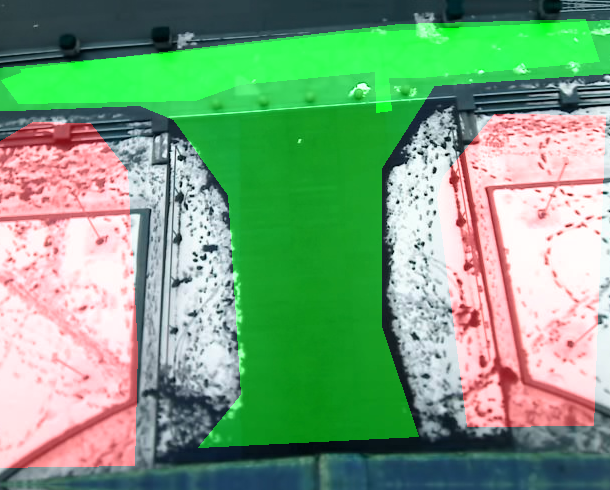} & 
\includegraphics[width=0.3\columnwidth]{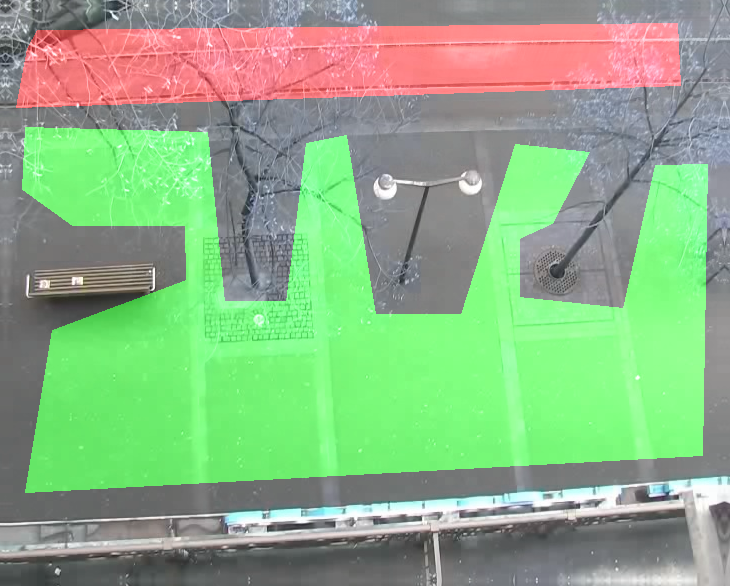} & 
\includegraphics[width=0.3\columnwidth]{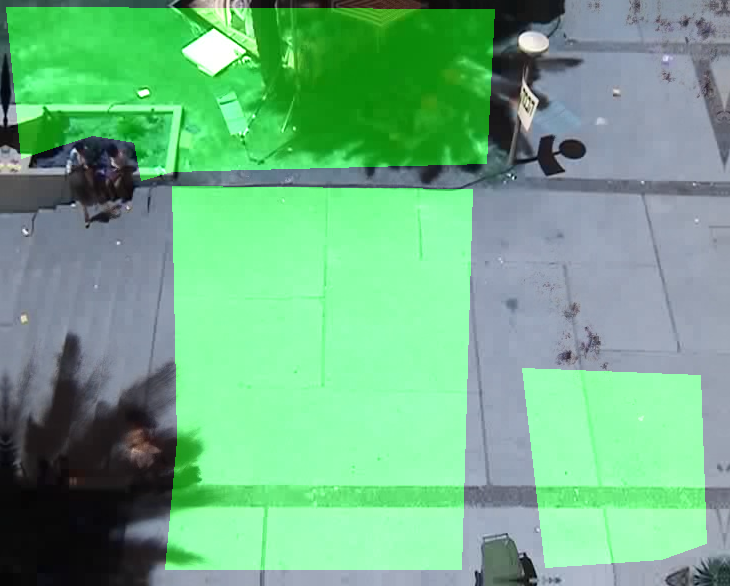} & 
\includegraphics[width=0.3\columnwidth]{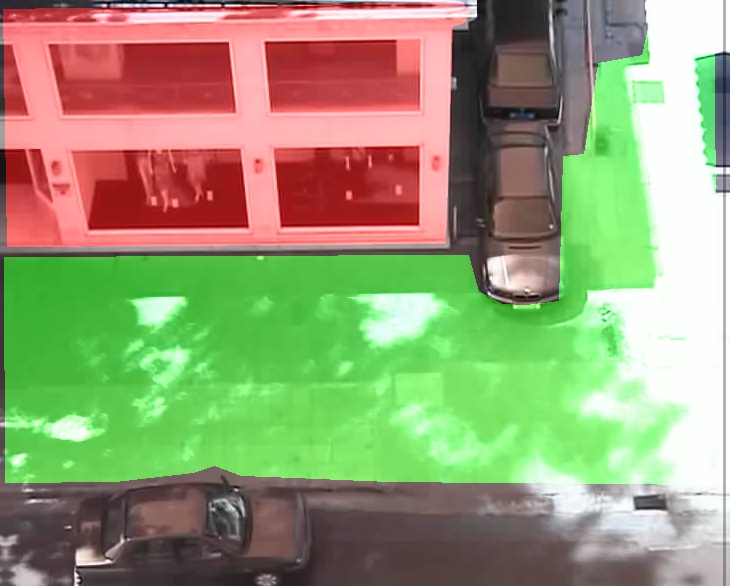} &
\includegraphics[width=0.3\columnwidth]{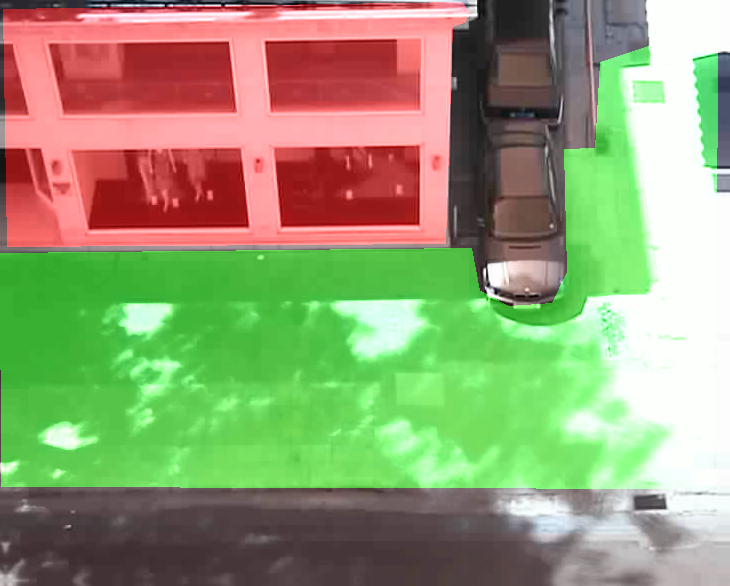}
\\
\raisebox{0.8cm}{\bf Obstacle} & 
\includegraphics[width=0.3\columnwidth]{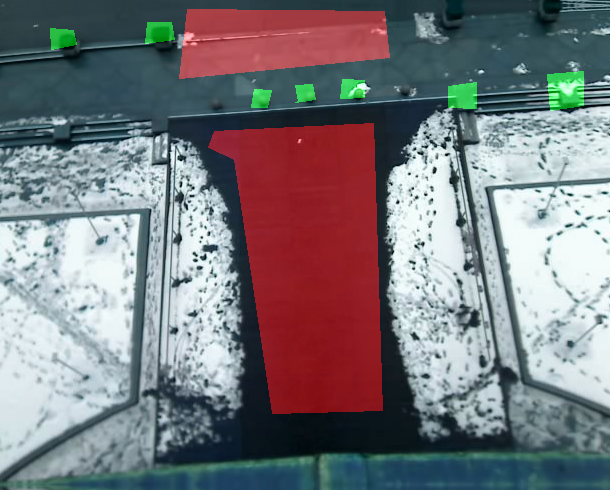} & 
\includegraphics[width=0.3\columnwidth]{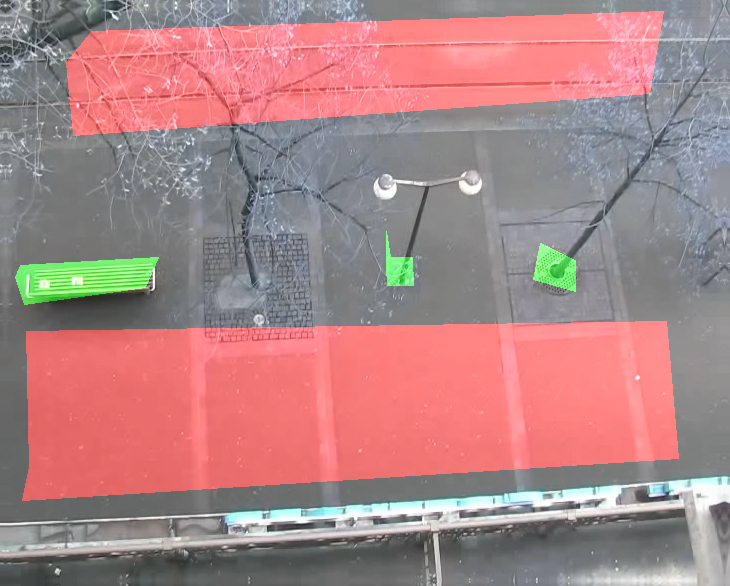} & 
\includegraphics[width=0.3\columnwidth]{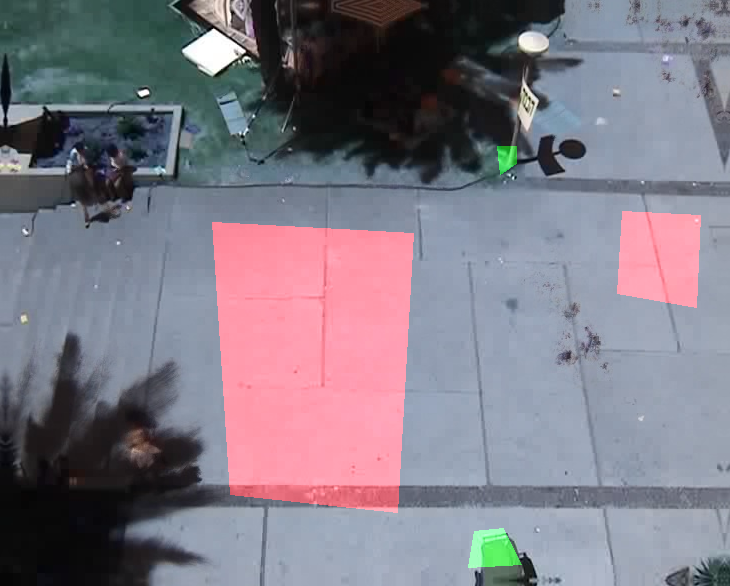} & 
\includegraphics[width=0.3\columnwidth]{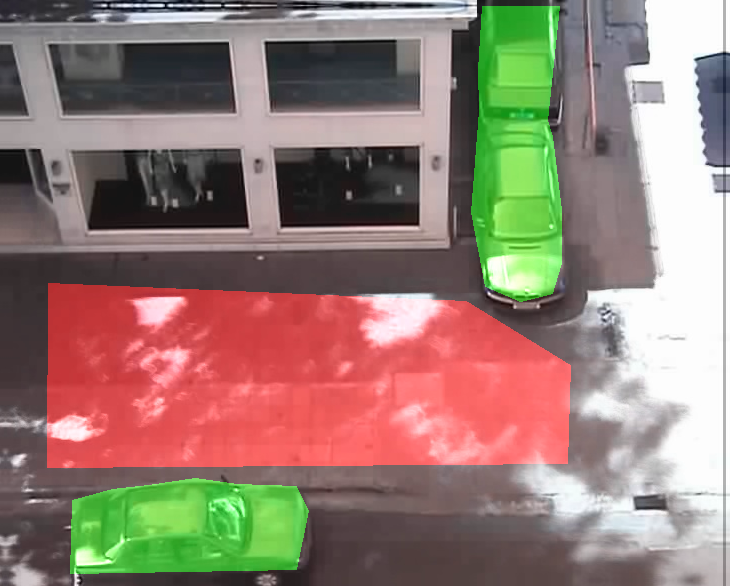} &
\includegraphics[width=0.3\columnwidth]{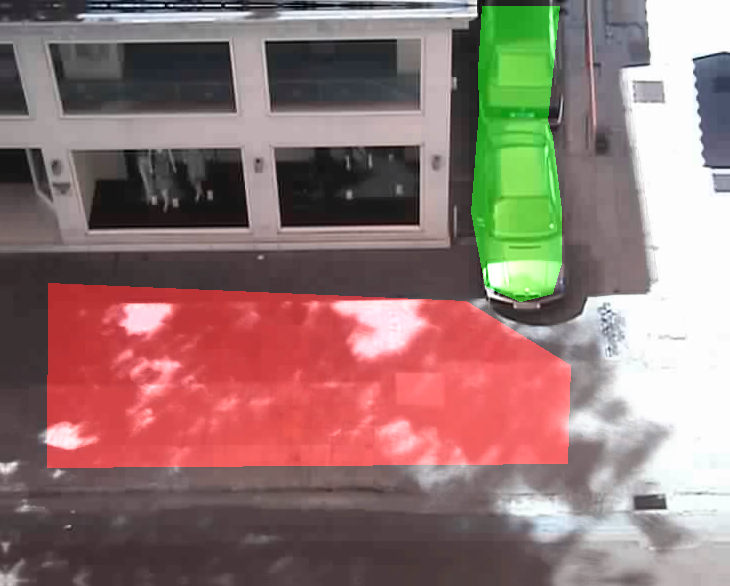}
\\
\end{tabular}
\caption{{\bf Annotated labels.} The labels are given in terms of positive examples (green) and negative examples (red) encoded as 1 and -1 respectively. They do not cover the full corresponding semantic area in every reference image. We demonstrate that the information encoded in those labels can still be used to learn segmentation and support prediction in unlabeled areas.} \label{tab:labels}
\end{table*}

\subsection{Sparsity}\label{sec:sparsity}
In order to ensure that the map is as simple as possible, we add a sparsity prior on the gradient of the latent map,
\[
\mcL_{\mathrm{sparsity}}(l) = \norm{\nabla M_l}_1\ .
\]
This is implemented using a finite differences approach computed by applying a convolution with the two predefined kernels
\[
\bbmat
0 &0 &0\\
\ve & -\ve & 0\\
0 & 0 & 0
\ebmat \text{ and }
\bbmat
0 &\ve &0\\
0 & -\ve & 0\\
0 & 0 & 0
\ebmat
\]
to the latent map. Then, we compute the norm by treating the resulting output as a vector. We refer the reader to \cite{elad2010sparse} and references therein for further details about sparse representations, and note that other image priors could have been used as well.

\subsection{Training} \label{sec:losses}
For a given scene, we initialize the context map and network with random weights picked from a Gaussian distribution. We alternate training between the trajectory generator network and the discriminator network. To further balance the training, the auxiliary map learning losses are additionally restricted to a set of randomly selected patches rather than the entire latent map.

\subsubsection{Generator Step}
 
To train the generator network, we initially do a forward pass by first selecting sections of the context map dependent on the scene and pedestrian coordinates in the batch. We input the coordinates and map sections into the network to predict the trajectories. We compute several losses with respect to the context map and the resulting trajectories to train the network.


During the training phase, we not only train the weights of the encoder as well as the Social-GAN modules but also the selected patches of the context map. Thus, the training process enforces location-specific information in the context map. The selected losses make sure that the information stored in the map informs the trajectory prediction network to make more accurate predictions and do so concisely in a way that captures the key components of the reference image.  

The total loss for the generator is thus obtained as
\begin{align*}
\mcL_{\mathrm{G}}(\fhY, \fY, l) &
= w_1 \cdot \mcL_{\mathrm{image}}(l) 
  + w_2 \cdot\mcL_{\mathrm{labels}}(l) \\
&\quad 
  + w_3 \cdot \mcL_{\mathrm{sparsity}}(l) 
  + w_4 \cdot \mcL_{\mathrm{score}}(\fhY)\\
& \quad  
  + w_5 \cdot \mcL_{\mathrm{traj}}(\fhY, \fY)\ .
\end{align*}
 
\subsubsection{Discriminator Step}
The descriminator is trained to provide $\mcL_{\mathrm{score}}$ in the same way as in Social GAN. We first pass a partial pedestrian trajectory to the trajectory generator to get a predicted full trajectory. We then pass the predicted trajectory and true trajectory to the discriminator network to obtain scores for the two trajectories.  While we did not modify the loss of the discriminator, we adapted the discriminator's first \ac{lstm} cell in the same way as in the generator. However, the discriminator does not have an own map encoder module and gets merely read-only access to the generator's maps, i.e. they are not changed during discriminator training. Letting the discriminator see the maps without modifying them allows for better convergence while not violating the \ac{gan}'s theoretical equilibrium properties.

\section{EVALUATION}
In our evaluation, we aim to capture the effects of using context maps on trajectory prediction accuracy. By explicitly discussing dataset imbalances and different scales, we demonstrate the utility of learned maps in the presence of a diverse set of data. In what follows we first introduce the baselines (Sec.~\ref{sec:baselines}) and datasets (Sec.~\ref{sec:datasets}), followed by the implementation details (Sec.~\ref{sec:implementation}) of our evaluation, and finally a discussion of our results (Sec.~\ref{sec:results}).

\begin{table*}[t]
\centering
\begin{tabular}{l|cccccc}
{\bf Sequence} & 
{\bf Linear} & 
{\bf S-GAN} & 
{\bf S-GAN-P} & 
{\bf Ours} & 
{\bf Ours no pooling} &
{\bf Ours no labels} 
\\ \hline 
ETH & 
16.33 / 35.09 & 
38.25 / 67.35 & 
44.62 / 81.96 & 
17.64 / 34.20 & 
{\bf 14.63} / {\bf 26.83} &
15.77 / 28.89 
\\
HOTEL & 
20.81 / 44.68 & 
23.52 / 39.57 & 
25.32 / 42.41 & 
19.12 / 34.62 & 
{\bf 18.79} / {\bf 33.77} &
20.90 / 38.20
\\
ZARA1 & 
21.44 / 49.16 & 
28.60 / 50.08 & 
30.87 / 53.45 & 
{\bf 17.79} / {\bf 34.54} & 
17.99 / 35.69 &
24.37 / 48.63
\\
ZARA2 & 
14.64 / 34.32 & 
19.48 / 34.56 & 
19.21 / 33.83 & 
13.43 / {\bf 26.20} & 
{\bf 13.32} / 26.40 &
15.88 / 31.09
\\
STUDENTS3 & 
30.86 / 71.38 & 
28.95 / 53.87 & 
29.55 / 54.81 & 
{\bf 22.02} / {\bf 43.84} & 
22.75 / 45.94 &
23.13 / 45.82
\\ \hline
\textbf{Average} &
20.81 / 46.93 &
27.76 / 49.09 &
29.91 / 53.29 &
18.00 / 34.68 &
\textbf{17.50} / \textbf{33.72} &
20.01 / 38.53
\end{tabular}
\caption{{\bf Prediction Results} given in terms of ADE (left) and FDE (right) in pixels for each sequence individually and a (non-weighted) average. Use of context maps outperforms approaches that purely predict from trajectory data. Particularly datasets underrepresented in the training data (in our case ETH) stand to benefit from the use of location-specific maps.} \label{tab:results}
\end{table*}

\subsection{Baselines}\label{sec:baselines}
We evaluate the proposed approach along the following baselines including several variants of our own model to understand the individual contribution of each component:

\subsubsection{Linear} A simple Kalman filter (we used pykalman 0.9.5) running a constant acceleration model where the initial state covariance, the model covariance, and the observation covariance were estimated using an expectation maximization method for each trajectory individually. 

\subsubsection{S-GAN-P} We compare our approach against the full S-GAN model including  their proposed pooling module. We use the S-GAN predictor (with the modifications outlined above) in our training pipeline and thus, this serves at the same time as an ablation study for the use of context maps. 
\subsubsection{S-GAN} This baseline is basically the same as S-GAN-P with the only difference being the removal of the pooling module. That is, we train a simple LSTM-based GAN for trajectory prediction.
\subsubsection{Ours} Our full model involving all loss terms described above.
\subsubsection{Ours no pooling} Our full model without the pooling module in the generator to evaluate the relative contribution of pooling. 
\subsubsection{Ours w.o. labels} Our full model without semantic labels in order to evaluate if weak semantic supervision helps prediction.

\subsection{Datasets}\label{sec:datasets}
The evaluation of our model requires datasets with trajectory data of multiple trajectories in the same location and a corresponding reference image of the scenery. Unfortunately, no such autonomous vehicle data is publicly available yet. Thus, we use datasets based on a static tracker and evaluate our work on the ETH and UCY standard benchmark datasets.

{\bf ETH Dataset~\cite{Pellegrini2009}} The ETH dataset, also known as BIWI Walking Pedestrian dataset, is a collection of two sequences (ETH and HOTEL) recorded around ETH Zurich. For both sequences, it contains pedestrian positions and velocities in meters and a video recording. Furthermore, it provides homography matrices for transforming the data into image coordinates. For dataset compatibility and to capture if the maps help with different scales, we use pixel coordinates for this dataset.

{\bf UCY Dataset~\cite{Lerner2007}.} The UCY "Crowds-by-Example" dataset contains several sequences with annotated pixel location trajectories. Not all of these sequences also have a corresponding video recording or the viewpoint of the recording, is not suitable. Thus, we make only use of the ZARA1, ZARA2, and STUDENTS3 sequences.

We process each dataset by stepping through sequences of frames and taking subsequences of size $18$ (with an observation sequence length of $O=10$ and an prediction sequence length of $P=8$). We only include pedestrians that appear in the full sequence of sampled frames. Partial pedestrian sequences are dropped. Since we use a sliding window approach, taking multiple (only partially overlapping) subsequences from each trajectory, we significantly increase the size and variety of our dataset. 

UCY is annotated at a rate of ten times that of ETH, so we only take every tenth frame of UCY to ensure that there are no large discrepancies in temporal scaling between the generated sequences from the two datasets. At the same time, we use each datapoint in both datasets as a potential starting point for a sequence. This data augmentation measure creates several partially overlapping sequences with more datapoints from UCY than ETH. We do not compensate for this in order to evaluate our hypothesis that context maps can better account for imbalance by properly learning location-specific information.

Once the datasets are processed, we split $10\%$ of the data in each scene into a validation set and $30\%$ of the data into a test set. The remaining $60\%$ are used for training. It is important to note that these percentages are taken from each scene, not from the overall pool. This ensures that there is a good representation of each scene in each of the splits. 
Furthermore, the splits were made on a temporal basis making sure that no two trajectories from the same point in time end up in different splits while at the same time covering most of the locations in training for context map learning.
In the original S-GAN work, different datasets were used for training and evaluation. To ensure a fair comparison, our approach uses the same data regime for our proposed method and the baselines. We provided semantic label annotations of the reference image for walkable areas and obstacles, i.e. $\mcT=\{\mathrm{walkable}, \mathrm{obstacle}\}$. They are visualized in Table~\ref{tab:labels}. 

\subsection{Implementation Details}\label{sec:implementation}
For the loss weights during generator training of our models, we use $w_1 = 0.05$, $w_2 = 0.05$, $w_3 = 0.5$, $w_4=1$, $w_5=0.1$. We use a batch size of $32$ and we train the model over $200$ epochs for convergence. For the training process, we alternate updates to the discriminator and the generator per batch using an Adam optimizer for both with a learning rate of 5e-3. 
For map patches, a size of $H_{patch} = W_{patch} = 10\mathrm{px}$ is used. This hyperparameter depends, in practice, on the dataset and resolution of the reference image in practice. 
For the $\mathrm{MapEncoder}_R$, we used the convolutional layers of a ResNet18 model and append a 3-layer convolutional head with $[10, 10, 2]$ output channels (and quadratic kernels of size $[1, 3, 3]$) respectively. For the semantic labels, prediction, and reconstruction encoders we have used nonlinear (ReLU) convolutional modules: feature count of $[5,5]$ and kernel size of $[5\times5,5\times5]$ for $\mathrm{MapDecoder}_S$, $[7,5,10]$ feature sizes and pixelwise convolutions for $\mathrm{MapDecoder}_P$, and a single 5-feature pixelwise operator for $\mathrm{MapDecoder}_R$.

Once training is done, we use the model that had the best performance on the validation set for reporting results on the test set. For all S-GAN models, including our variant based on latent maps, the evaluation is based on letting the generator draw one sample. All decoder networks have been implemented as \acp{cnn}. The full implementation will be made available upon acceptance of this work.

%
%
\begin{figure*}
	\centering
	\begin{subfigure}[t]{\textwidth}
		\centering
		\includegraphics[width=0.32\textwidth]{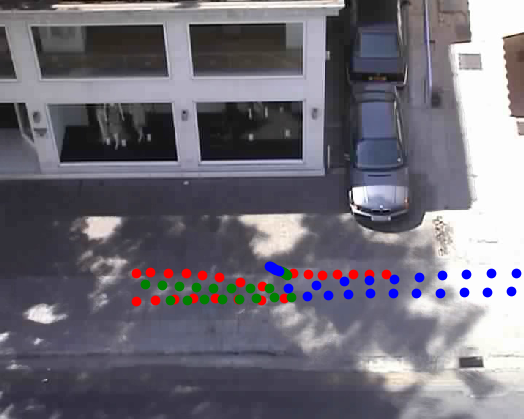}%
		~
		\includegraphics[width=0.32\textwidth]{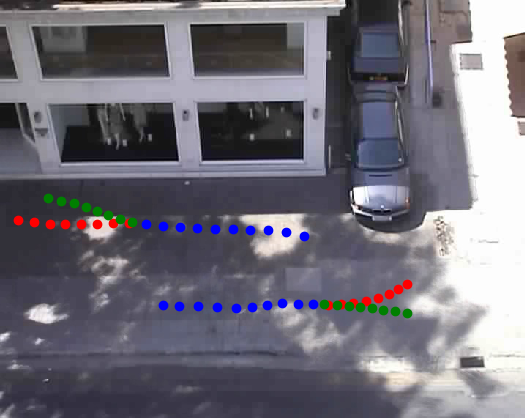}%
		~
		\includegraphics[width=0.32\textwidth]{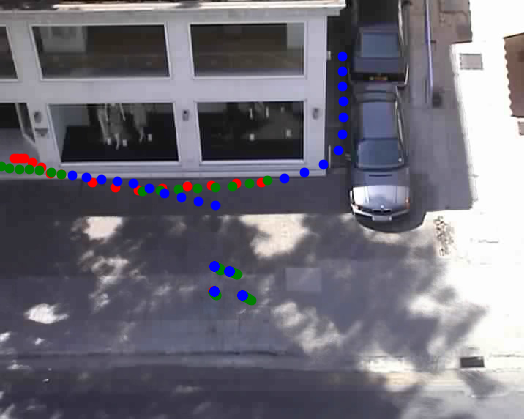}
		\caption{Examples from the ZARA2 sequence.}
	\end{subfigure}
	\begin{subfigure}[t]{\textwidth}
		\centering
		\includegraphics[width=0.32\textwidth]{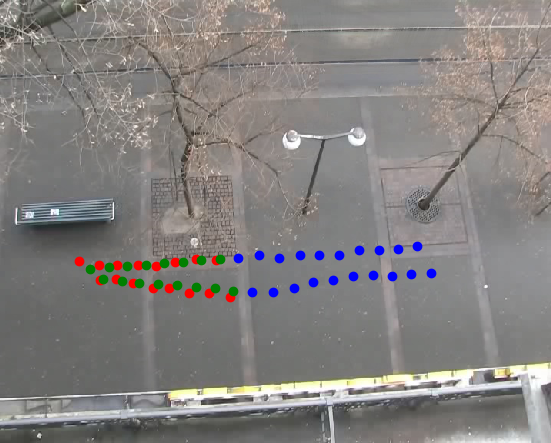}%
	    ~ 
		\includegraphics[width=0.32\textwidth]{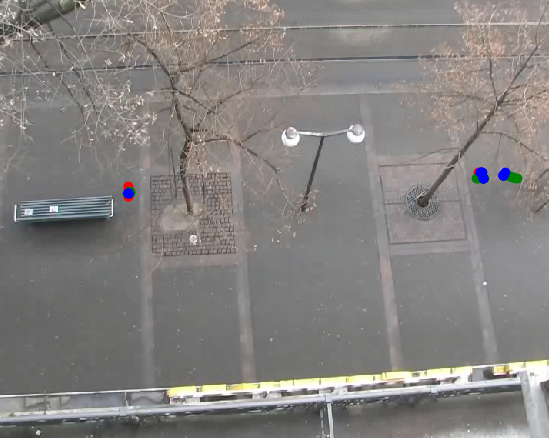}%
	    ~
		\includegraphics[width=0.32\textwidth]{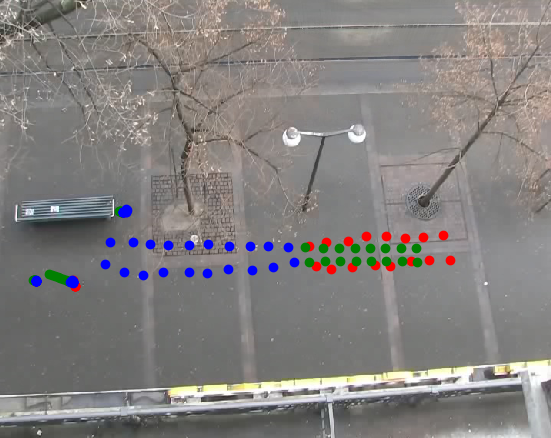}
		\caption{Examples from the HOTEL sequence.}
	\end{subfigure}%
	
	\caption{{\bf Example Predictions.} Several selected example trajectories (red) from different scenes with corresponding predictions by the network (green) and the observed past trajectories (blue) that were used for generating the predictions.}\label{fig:example_predictions}
\end{figure*}

\subsection{Results}\label{sec:results}
Similarly to \cite{Gupta2018}, we evaluate the performance of our model on trajectory prediction using two metrics: Average Displacement Error (ADE) and Final Displacement Error (FDE) which are also known as Mean L2 Error (ML2) and Final L2 error (FL2)~\cite{Sadeghian2019}.
The ADE, given as
\begin{equation*}
\mathrm{ADE} 
= \frac{1}{N \cdot P}\sum_{i=1}^{N} \sum_{t=1}^{P} 
  \norm{\fy_{i,t}-\fhy_{i,t}}\ ,
\end{equation*}
averages the error between every position in the prediction and the ground truth. The FDE, given as
\begin{equation*}
\mathrm{FDE} 
= \frac{1}{N}\sum_{i=1}^{N}
  \norm{\fy_{i,P}-\fhy_{i,P}}\ ,
\end{equation*}
denotes the error at the last predicted position. Both metrics are averaged over the entire test set.

%
%
The results are visualized in Table~\ref{tab:results}. Overall, use of the newly proposed context maps strongly improves the results compared to merely using different variants of S-GAN. This is mainly due to the fact that it becomes easier to tailor the prediction to the data and the environment. The qualitative differences to the original S-GAN results are mainly due to the fact that we learn directly on pixel space and have a different data preparation and augmentation process. Particularly the absence of a homography unifying the scale of the trajectories makes it more challenging to properly predict trajectories at different scales without location-specific memory. 

%
%
Our work confirms that the contribution of the pooling module is minor (for the prediction task) compared to storing context. This, however, is partially due to the datasets not containing enough interactions such as near collisions. Furthermore, the strong improvements on the ETH sequence compared to not using context maps confirms our hypothesis that learned maps are particularly beneficial in situations with dataset imbalances and changing viewpoints.

\subsection{Qualitative Examples}
Several example trajectories from the ZARA2 and HOTEL sequences are visualized in Figure~\ref{fig:example_predictions}. These examples also involve some failure cases: In the ZARA2 sequence, the network may assume that pedestrians are likely to enter the store when close to the entrance as this is a frequently observed behaviour at that location. Overall, however, this ability of the proposed predictor to capture typical trajectories and behaviours at a given location results in a better prediction accuracy. At the same time, the auxiliary loss terms help to avoid over-fitting during training.

\section{CONCLUCSION}
In this work, we presented Deep Context Maps, a map learning approach for agent trajectory forecasting. We demonstrated how this approach can be integrated into a state-of-the-art predictor and that it achieves significant improvements on the agent trajectory forecasting task. Overall, maps promise to avoid over-fitting to the location of the training set in deep-learning based inference tasks for autonomous systems that are deployed in a big variety of diverse environments.

\bibliographystyle{packages/IEEEtran}
\bibliography{literature}

\end{document}